\documentclass{article}

\usepackage[final]{neurips_2024}

\usepackage[utf8]{inputenc} % allow utf-8 input
\usepackage[T1]{fontenc}    % use 8-bit T1 fonts
\usepackage{hyperref}       % hyperlinks
\usepackage{url}            % simple URL typesetting
\usepackage{booktabs}       % professional-quality tables
\usepackage{amsfonts}       % blackboard math symbols
\usepackage{nicefrac}       % compact symbols for 1/2, etc.
\usepackage{microtype}      % microtypography
\usepackage{xcolor}         % colors
\usepackage{bm}             % bold maths
\usepackage{graphicx}       % figures
\usepackage{caption}        % custom captions
\usepackage{amsmath}        % advanced math equations
\usepackage{amssymb}        % advanced math symbols
\usepackage{microtype}      % advanced word cutting and justification
\usepackage{fontawesome}

\title{Integrating Physics and Topology in Neural Networks for Learning Rigid Body Dynamics}

\author{%
  Amaury Wei\\
  EPFL - IMOS Laboratory\\
  CH-1015 Lausanne\\
  Switzerland\\
  \texttt{amaury.wei@epfl.ch} \\
  \And
  Olga Fink\textsuperscript{\faEnvelope}\\
  EPFL - IMOS Laboratory\\
  CH-1015 Lausanne\\
  Switzerland \\
  \texttt{olga.fink@epfl.ch} \\
}

\begin{document}

\maketitle

%%%%%%%%%%%%%%%%%%%%%%%%%%%%%%%%%%%%%%%%%%%%%%%%%%%%%%%%%%%%
\begin{abstract}

Rigid body interactions are fundamental to numerous scientific disciplines, but remain challenging to simulate due to their abrupt nonlinear nature and sensitivity to complex, often unknown environmental factors. These challenges call for adaptable learning-based methods capable of capturing complex interactions beyond explicit physical models and simulations. While graph neural networks can handle simple scenarios, they struggle with complex scenes and long-term predictions.\linebreak We introduce a novel framework for modeling rigid body dynamics and learning collision interactions, addressing key limitations of existing graph-based methods. Our approach extends the traditional representation of meshes by incorporating higher-order topology complexes, offering a physically consistent representation. Additionally, we propose a physics-informed message-passing neural architecture, embedding physical laws directly in the model. Our method demonstrates superior accuracy, even during long rollouts, and exhibits strong generalization to unseen scenarios. Importantly, this work addresses the challenge of multi-entity dynamic interactions, with applications spanning diverse scientific and engineering domains.

\end{abstract}

%%%%%%%%%%%%%%%%%%%%%%%%%%%%%%%%%%%%%%%%%%%%%%%%%%%%%%%%%%%%
\section{Introduction}

% Paragraph 1: Introduce motivation + Describe the type of problem tackled by our method

%% Motivation + why simulators are not applicable

Modeling dynamic interactions between rigid bodies is an essential challenge across many scientific and engineering disciplines, including biomechanics, robotics, aerospace engineering, and virtual reality. It is not only essential for understanding complex systems and their dynamics, but it also supports the design of advanced engineered systems with counterfactual reasoning\textemdash enabling hypotheses to be tested in realistic virtual environments. When all physical parameters of a system are precisely known, traditional general-purpose simulators, such as MuJoCo \cite{mujoco}, Bullet \cite{bullet}, ODE \cite{ode}, or PhysX \cite{physx}, can be used to generate plausible predictions about system behavior.
However, in many real-world scenarios, key aspects of the environment remain unknown\textemdash such as current wind conditions, pressure, temperature, and gravity\textemdash or are too complex to model explicitly, including surface texture and wetness, material damping, and fatigue. This challenge is particularly relevant for autonomous systems, which must swiftly react to dynamic objects in their surroundings while operating with limited knowledge of the environment\textemdash for example, unfamiliar locations, unpredictable weather conditions, or newly encountered obstacles. Consequently, in situations where numerous aspects of a scene are uncertain, learning-based approaches offer a more practical and effective solution, allowing systems to implicitly learn missing parameters from past observations. % These approaches enable systems to adapt and respond to unforeseen complexities, enhancing their robustness and versatility in real-world applications.

\begin{figure}[t]
    \centering
    % \vspace{-0.8cm}
    \includegraphics[width=0.95\textwidth]{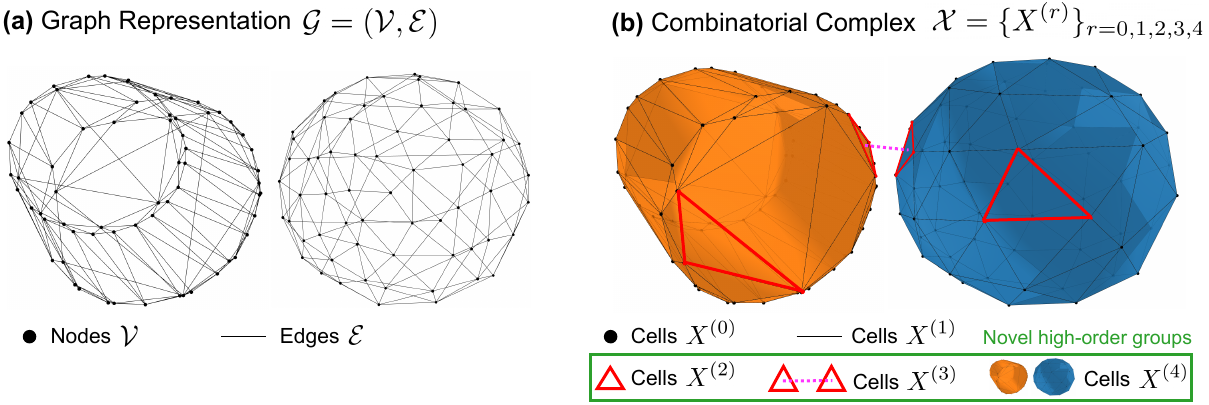}
    \captionsetup{labelfont=bf}
    \caption{
        \textbf{Comparison between graph and combinatorial complex representation.} (\textbf{a}) Standard graph representation $\mathcal{G} = (\mathcal{V}, \mathcal{E})$ used in \cite{meshgraphnet,fignet}; (\textbf{b}) Our  proposed Combinatorial Complex representation $\mathcal{X}$. While the graph $\mathcal{G}$ only supports nodes and edges, the combinatorial complex $\mathcal{X}$ enables us to define a hierarchy of higher-order groups of nodes $X^{(r)}$ (e.g., mesh triangles $X^{(2)}$, collision contacts $X^{(3)}$, and objects $X^{(4)}$) and assign distinct features to them. Only four $X^{(2)}$ cells and one $X^{(3)}$ cell are shown in (b) to preserve visibility.
    }
    \label{fig:mesh_representation}
    \vspace{-0.3cm}
\end{figure}

Furthermore, learning directly from data can bridge the often substantial simulation-to-reality gap. Traditional simulation tools are limited in their accuracy due to the inherently complex nature of rigid body interactions. They necessitate explicit computation of forces, including contact and friction forces, which are notoriously difficult to model accurately \cite{challenges_stiff_contact}. In contrast, data-driven approaches can implicitly capture intricate and subtle dynamics\textemdash such as the fine granularity of object surfaces that significantly affects collisions\textemdash thereby offering a more comprehensive and realistic model for physical interactions.

% Paragraph 2: How people have been solving the task until now (related work)

%% Vision-based approaches

To effectively learn rigid body dynamics, numerous approaches have been proposed, typically falling into one of two categories: vision-based or graph-based.
Focusing on vision-based methods, recent advancements \cite{plato,savi,savi++,slot_transformer,slotformer} utilize unsupervised object-centric frameworks \cite{monet,slot_attention} to address the problem. These models leverage multiple attention mechanisms \cite{attention} to isolate individual objects in images and predict their dynamics based on past and current observations. While these approaches achieve high accuracy in simple scenarios, they are constrained by the necessity for a predefined, fixed number of objects and suffer from limited interpretability. Moreover, vision-based methods, which rely on 2D representations of an inherently 3D world, often struggle to handle complex scenarios. They are particularly sensitive to occlusions, which are frequent in scenes with multiple objects. Despite recent efforts to enhance their reasoning capabilities \cite{aloe,comphy,savi++} and improve the tracking of hidden objects, vision-based methods suffer from poor accuracy in long rollouts in complex scenarios \cite{physion}. This limitation arises because images provide only partial and localized representations of the environment, making it challenging to maintain consistent and accurate predictions over extended periods.

%% Graph-based approaches

Graph Neural Networks (GNNs) \cite{gnn} have emerged as a powerful framework for modeling complex relational data by effectively capturing and propagating information across interconnected entities. This capability makes them well-suited for representing the intricate and dynamic interactions inherent in 3D environments. Unlike vision-based methods that rely on indirect feature extraction, GNNs provide a more accurate and direct representation of 3D interactions by explicitly modeling the underlying relational structure of the data. Within this framework, particle-based methods \cite{mrowca2018flexible,propnet,dpinet,graph_network_simulator,physics_consistent_particles,subequivariant_gnn,gns_collisions}, which represent each object as a collection of dense particles, have proven effective in simulating liquids, soft materials, simple rigid bodies \cite{graph_network_simulator}, and contact collisions \cite{gns_collisions}. However, these methods face significant scalability challenges because their computational cost increases quadratically with scene complexity (e.g., number of objects and collisions). Furthermore, their ability to accurately model collisions is constrained by the particle-based representation, which lacks physical consistency; real-world collisions occur between continuous surfaces, not individual particles.

Recently, graph-based methods that operate directly on surfaces have been developed  \cite{meshgraphnet,fignet}. These approaches transform the meshes of rigid bodies into graph structures composed of vertices and edges, which are then processed using message-passing architectures \cite{relational_inductive_biases}. Several of these methods introduce surface- or object-level representations \cite{mrowca2018flexible,meshgraphnet,fignet} using heuristics such as virtual nodes \cite{fignet} or hierarchical clustering \cite{mrowca2018flexible}. While such strategies improve upon particle-based models, they remain limited in how explicitly they encode the underlying physical and geometric structure. Although these methods introduce higher-level concepts, their internal computations still primarily rely on node-level message passing, which may limit their ability to capture emergent group-level dynamics\textemdash crucial in rigid body systems where nodes, surfaces, and objects are tightly coupled. Additionally, GNNs face inefficient information propagation: messages can only reach immediate neighbors (1-hop) per layer, requiring many layers to reach distant nodes. This, in turn, introduces issues like over-smoothing \cite{gnn_oversmoothing}, ultimately making training more challenging and reducing model performance.

To partially address those limitations, Allen et. al. \cite{fignet} proposed duplicating shared object properties across each node and introducing virtual points to facilitate long-range information transfer within objects. Nevertheless, the way neural messages propagate across the graph does not accurately reflect the underlying physical interactions. Specifically, although collisions are computed between mesh triangles, the subsequent message passing occurs in a traditional "node-to-node" fashion\textemdash propagating from the points of contact to all other nodes within the object. This approach is inconsistent with Newtonian physics governing inelastic collisions, where energy and momentum are transmitted centrally through the center of mass rather than being distributed solely through individual nodes.

% Paragraph 3: Link related work with topology and physics-informed biases

% How to improve the representation of objects for neural networks ... and how to better process interactions ...
% "Breakthrough" moment with Topological Deep Learning

Ensuring consistency between real-world phenomena and their representation in deep neural networks is essential for effectively capturing complex interactions. In the context of rigid body dynamics, this requires improved modeling  of rigid bodies and a refined processing of collisions. A promising direction involves leveraging higher-order groups of nodes within graph-based data \cite{Battiston2021,nature_simplicial,nature_higher_order}, which has found promising applications in domains like chemistry \cite{schnet,Fang2022,topologynet,schutt2021equivariant} and biology \cite{Pineda2023}. This approach, known as Topological Deep Learning \cite{topological_deep_learning,bodnar_2022,topological_deep_learning_survey}, has demonstrated promising results in structured data environments \cite{scn_trajectory_pred,cw_networks}. However, its application to geometric data remains predominantly limited to static scenarios, primarily focusing on classification tasks while overlooking temporal dynamics and multi-entity interactions. Moreover, existing topological neural networks typically do not incorporate physical knowledge into their message-passing mechanisms. 

\begin{figure}[t]
    \centering
    \includegraphics[width=1\textwidth]{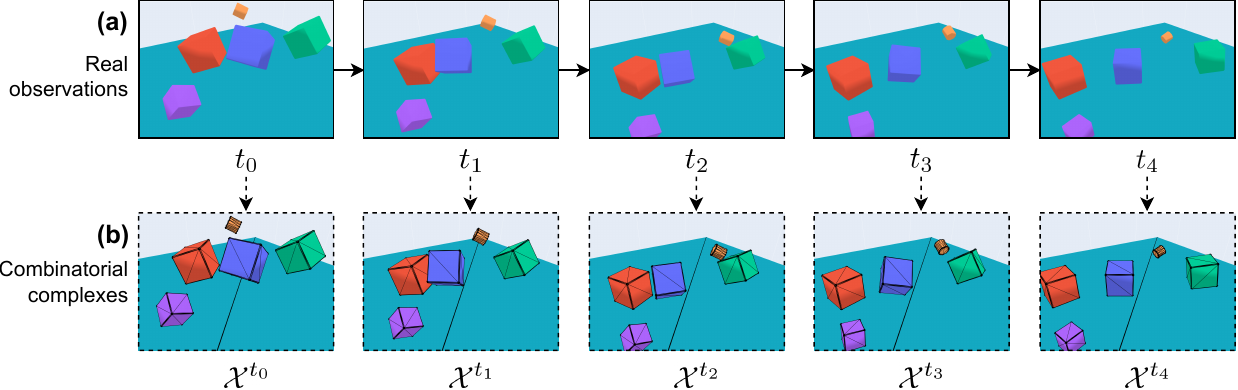}
    \captionsetup{labelfont=bf}
    \caption[Spatiotemporal combinatorial complexes]{
        \textbf{Spatiotemporal combinatorial complexes} (\textbf{a}) Real-world observations are represented by (\textbf{b}) spatiotemporal combinatorial complexes $\{\mathcal{X}^t\}$. The topology and features of each combinatorial complex evolve over time to accurately represent the environment. The illustrated sequence is taken from experiment 1 of the \texttt{MOVi-A} dataset.
    }
    \label{fig:spatio_temporal_ccs}
    \vspace{-0.3cm}
\end{figure}

% Paragraph 4: High-level introduction of the method + contributions

In this article, we propose "Higher-Order topological Physics-informed Network" (HOPNet), a novel and versatile learning-based message-passing framework designed to (1) model complex dynamic object geometries using higher-order spatiotemporal topological groups, and (2) learn rigid body dynamics through physics-informed topological neural architectures operating on these topological representations. Unlike prior approaches that heuristically approximate object structure, HOPNet formally encodes surfaces and objects using evolving topological structures that group relevant nodes and establish a clear hierarchy among nodes, mesh triangles, and objects. Learnable messages are exchanged between these higher-order groups along pathways specifically defined based on Newtonian principles, thereby integrating physics-based knowledge directly into the learning process. We evaluate our framework on tasks involving forward dynamics and collisions between rigid bodies in increasingly complex scenarios. HOPNet demonstrates high accuracy with complex shapes and multi-object collisions, successfully capturing highly nonlinear and abrupt dynamics while maintaining its performance over long rollouts. Additionally, the method exhibits strong generalization to both real-world geometries and counterfactual scenarios, indicating its effectiveness in learning the underlying fundamental physical laws. This counterfactual capability enables rigorous exploration of possible outcomes under various hypothetical conditions, reinforcing our framework's utility in diverse scientific and engineering domains. Finally, our framework's learning efficiency and robustness to hyperparameter selection further minimize the need for fine-tuning, supporting broad applicability and ease of deployment across different tasks.

%%%%%%%%%%%%%%%%%%%%%%%%%%%%%%%%%%%%%%%%%%%%%%%%%%%%%%%%%%%%
\section{Higher-Order topological Physics-informed Network (HOPNet)}

\subsection{Object meshes as spatiotemporal topological groups} \label{subsec:topology_groups}

In this work, we introduce HOPNet,  a deep learning framework designed to learn the dynamic interactions between multiple rigid bodies with diverse shapes and physical properties. By learning directly from observational data, HOPNet implicitly captures the dynamics of both single and multiple objects utilizing surfaces as inputs in a manner consistent with traditional rigid-body simulators.

Rigid-body simulations represent objects as meshes, where at a given timestep $t$, each mesh $M^t$ is defined by a set of node positions $\{\mathbf{x}_i\}_{i=1..N}$ and a set of faces $\{f_j\}_{j=1..F}$. The faces describe the connections between nodes, forming the surface of the mesh. Each triangular face $f_j$ is defined by three nodes $\{a,b,c\}$, where $a$, $b$, and $c$ are unique node indices within the face. While all meshes in our experiments utilize triangular faces, the proposed framework can generalize to faces with different numbers of nodes (e.g., tetrahedron, hexahedron, etc.).

\begin{figure}[t]
    \centering
    \vspace{-0.8cm}
    \includegraphics[width=1.0\textwidth]{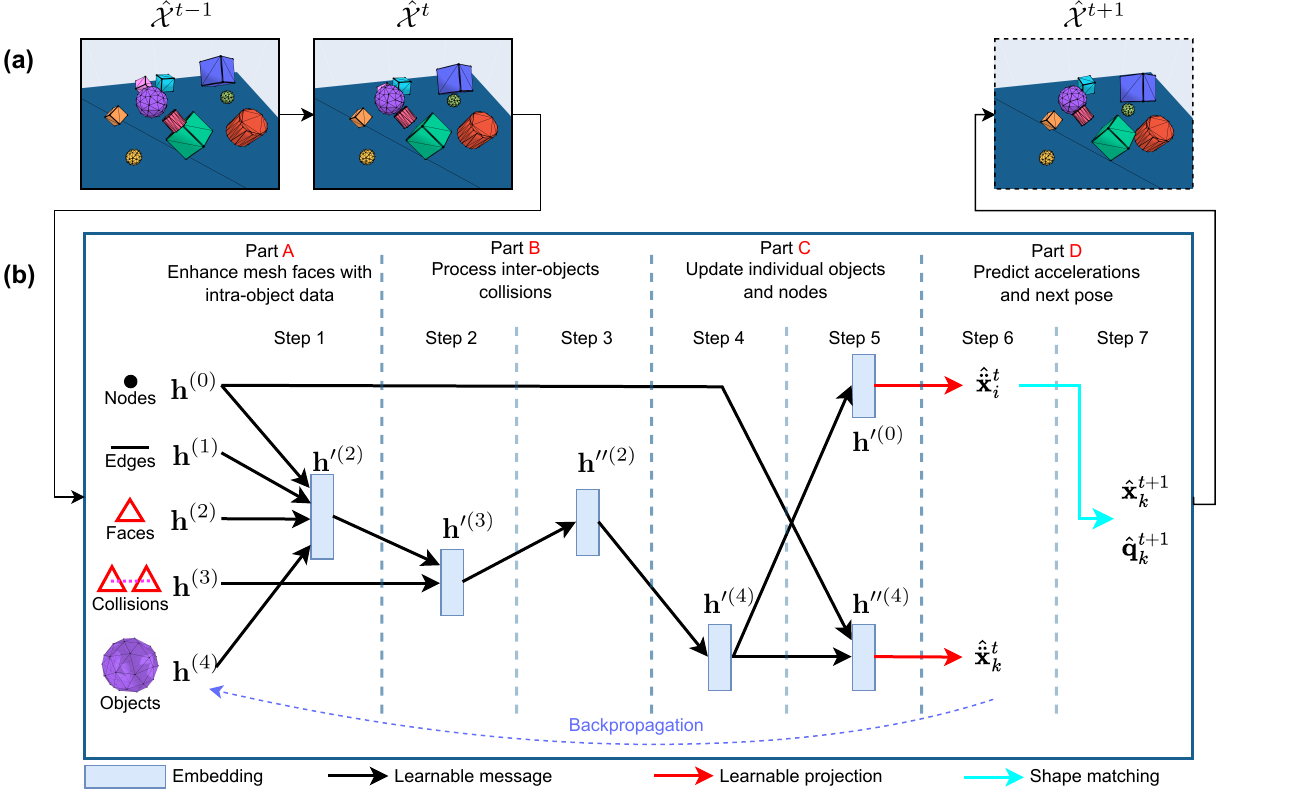}
    \captionsetup{labelfont={color=black,bf}}
    \caption[Overview of our method]{\textbf{Overview of our method.} (\textbf{a}) Autoregressive rollout approach; (\textbf{b}) Physics-informed message-passing strategy. Our sequential message-passing is inspired by Newtonian laws and tailored to process collisions. It first processes ongoing collisions (steps 1, 2, and 3) and then updates individual objects (steps 4 and 5). Finally, the per-node $\hat{\ddot{\mathbf{x}}}_i$ and per-object $\hat{\ddot{\mathbf{x}}}_{k}$ accelerations are computed (step 6) and the final object poses $(\text{positions } \hat{\mathbf{x}}^{t+1}_{k},\text{quaternionic orientations }\hat{\mathbf{q}}^{t+1}_{k})$ are obtained (step 7) with shape matching. The embedding vectors of each cell type $X^{(r)}$ of rank $r$ on a combinatorial complex $\mathcal{X}$ are noted $\mathbf{h}^{(r)}$. Other mathematical symbols are explained in "\hyperref[sec:methods]{Methods}" and summarized in Supplementary Information Section 1. The illustrated sequence is taken from experiment 2 of the \texttt{MOVi-A} dataset.}
    \label{fig:method_overview}
    \vspace{-0.4cm}
\end{figure}

At any given timestep $t$, the entire environment is represented by a single global mesh $M^t$ that may contain one or multiple objects. In previous works \cite{meshgraphnet,fignet}, this global mesh $M^t$ was converted into a standard graph $\mathcal{G} = (\mathcal{V}, \mathcal{E})$, where vertices $\mathcal{V} = \{v_i\}$ represent the mesh nodes and edges $\mathcal{E} = \{e_{v_s\rightarrow v_r}\}$ are bidirectionally created between the nodes of each face (as illustrated in Figure \hyperref[fig:mesh_representation]{1a}). However, this approach discards critical information about the scene: (1) cohesion of each individual object: the integrity of each individual object within the global mesh is not preserved, and (2) the consistency of surface triangles: the uniformity and structural integrity of the mesh surfaces are lost. In this uniform representation, all nodes are treated as identical entities, and objects are only distinguishable by the absence of connections between their respective nodes. Consequently, by losing most of the structural information and retaining only the lowest-level connectivity, the representation is reduced to simple node-to-node connections, thereby oversimplifying the true geometry.

To overcome the inherent limitations of previous approaches, our framework introduces a novel representation that preserves both the cohesion of individual objects and the consistency of mesh surfaces. This is achieved by defining meshes as higher-order spatiotemporal topological groups. We propose to represent the global mesh $M^t$ as a Combinatorial Complex (CC)\textemdash a higher-order domain that allows the definition of groups of nodes (called "cells") and introduces a hierarchical structure between them (called "rank" and denoted $r$). Unlike other methods \cite{meshgraphnet,fignet,gns_collisions}, our representation allows for an explicit, physics-consistent encoding of both mesh triangles and entire objects of the global mesh $M^t$, enabling a more faithful modeling of real-world dynamics. Our proposed representation $\mathcal{X}$ uses cells of five different types, as illustrated in Figure \hyperref[fig:mesh_representation]{1b}: nodes $X^{(0)}$, edges $X^{(1)}$, mesh triangles $X^{(2)}$, collision contacts $X^{(3)}$, and objects $X^{(4)}$. Each cell $X^{(r)}$ is assigned specific features $\mathbf{h}^{(r)}$\textemdash such as velocity for nodes or mass for objects\textemdash unlike traditional graph structures, where only nodes and edges are associated with features. This enriched representation captures both geometric and physical properties, improving the model's ability to handle complex interactions between rigid bodies.

% TODO: Mention that combinatorial complexes have been used for chemistry but never for spatiotemporal

Finally, we represent the evolution of multi-body environments over time using a sequence of spatiotemporal combinatorial complexes $\{\mathcal{X}^{t_0}, \mathcal{X}^{t_1}, \mathcal{X}^{t_2}, ...\}$, illustrated in Figure \ref{fig:spatio_temporal_ccs}. Each object in $\mathcal{X}^t$ has a specific position and rotation which evolves over time. Our method estimates the next state of the system $\hat{\mathcal{X}}^{t+1}$ from a history of previous states $\{\mathcal{X}^{t-1}, \mathcal{X}^{t}\}$. This enables trajectory rollouts by iteratively feeding the model with its previous predictions, generating a sequence such as $\{\mathcal{X}^{t_0}, \mathcal{X}^{t_1}, \hat{\mathcal{X}}^{t_2}, \hat{\mathcal{X}}^{t_3}, ..., \hat{\mathcal{X}}^{t_k}\}$. This autoregressive approach is illustrated in Figure \hyperref[fig:method_overview]{3a}.

\begin{figure}[t]
    \centering
    \vspace{-0.8cm}
    \includegraphics[width=1\textwidth]{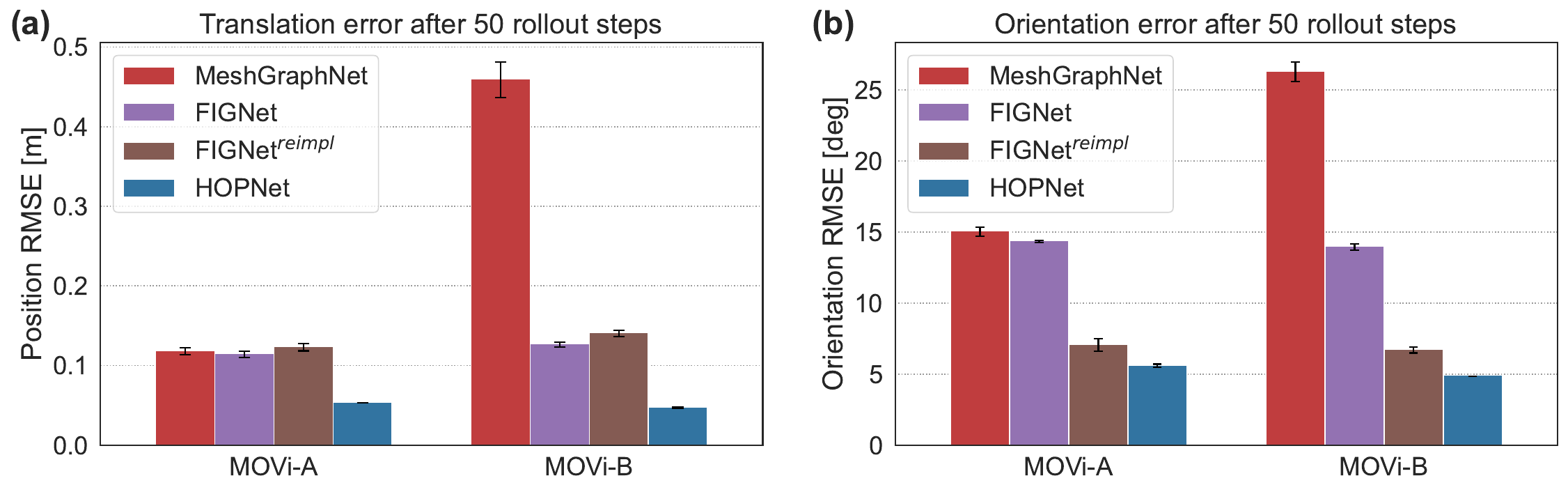}
    \captionsetup{labelfont={bf}}
    \caption{
        \textbf{Autoregressive rollout performance on benchmark datasets.} (\textbf{a}) Translation and \linebreak(\textbf{b}) orientation errors after a rollout horizon of $T=50$ on all dynamic objects in the scene. Error bars indicate the mean and standard deviation across three independent random seeds. Comparisons with FIGNet \cite{fignet} and MeshGraphNet \cite{meshgraphnet} are based on the results reported in \cite{fignet}.
    }
    \label{fig:standard_results}
    \vspace{-0.5cm}
\end{figure}

\subsection{Physics-informed learnable messages} \label{sec:physics_learnable_messages}

By introducing higher-order groups within combinatorial complexes, we establish physically-motivated pathways for information flow between cells. Unlike standard message-passing neural architectures \cite{relational_inductive_biases} used on graphs\textemdash where information is typically exchanged only between individual nodes through directed edges\textemdash our framework allows learnable messages can be transmitted between any two cells that share at least one common node. This extension enables efficient information flow not only between nodes, but also between mesh elements and object-level groups, effectively modeling more complex relationships and interactions. Additionally, it supports long-range information transfer without requiring the multiple layers of message-passing typically needed in traditional GNNs. Consequently, this enhances the expressiveness, learning efficiency, and accuracy of the learned dynamics.

The incorporation of higher-order groups provides greater flexibility in exchanging information between rigid bodies and their surfaces. Rather than allowing information to flow indiscriminately among all cells, we embed physics-informed inductive biases to define precise pathways for message-passing within the combinatorial complexes. This innovative approach offers several key benefits: (1) Reduced computational costs and model complexity: by restricting message-passing to targeted pathways, we minimize unnecessary computations and streamline the model architecture;\linebreak(2) Enhanced learning efficiency: physically-grounded pathways guide the neural network, accelerating the learning process and improving convergence; and (3) Improved model explainability: aligning message flows with physical principles enhances the interpretability of the model, making it easier to understand and trust its predictions.

To simulate rigid body dynamics, we incorporate Newtonian laws to orchestrate a sequential flow of learnable messages. These messages are computed using MultiLayer Perceptrons (MLPs) with learnable weights, which are optimized via gradient descent to capture the underlying physics of motion and collisions. To effectively capture the abrupt and highly nonlinear dynamics of collisions, we consider two distinct scenarios: (1) Independent evolution of objects: here, an object's trajectory is only influenced by external forces. Using the principles of conservation of momentum and energy,  we compute its acceleration solely based on its current state; and (2) Colliding objects: in this scenario,  collisions result in the exchange and loss of energy between colliding objects. This requires an additional initial step to process the collision, transfer energy, and update each object's state before computing their respective accelerations.

% By addressing these scenarios, our framework effectively models both the independent and interactive dynamics of rigid bodies, capturing highly nonlinear and abrupt changes while maintaining robust performance over extended rollouts. Additionally, the method exhibits strong generalization to unseen and counterfactual scenarios, demonstrating its capability to learn and apply fundamental physical laws accurately.

We formalize these fundamental principles into a unified, sequential, physics-informed higher-order message-passing framework for Combinatorial Complexes (CCs), as illustrated in Figure \hyperref[fig:method_overview]{3b}. This novel approach is specifically designed to capture abrupt nonlinear dynamics, thereby enhancing the modeling of rigid-body dynamics. First, the features of each triangular face, $\mathbf{h}^{(2)}$, are enriched by incorporating the features of their constituent nodes, $\mathbf{h}^{(0)}$, and the corresponding high-level object features, $\mathbf{h}^{(4)}$ (step 1). In the event of collisions, changes in energy and momentum are computed using collision contacts $X^{(3)}$ (step 2). These collision contacts $X^{(3)}$ are defined based on inter-object distance: when objects come within a certain radius $d_c$ of each other, a collision contact cell is created. This mechanism follows a standard practice in neural physics simulators \cite{dpinet,propnet,fignet}, where connectivity is dynamically updated based on spatial proximity. The computed changes in energy and momentum are then propagated back to the colliding higher-order objects $\mathbf{h}^{(4)}$ (steps 3 and 4). Finally, predictions at both object and node levels are generated based on the current internal states of each object, $\mathbf{h}^{(4)}$, and its nodes $\mathbf{h}^{(0)}$ (steps 5 and 6).

\begin{figure}[t]
    % \vspace{-1cm}
    \centering
    \includegraphics[width=1\textwidth]{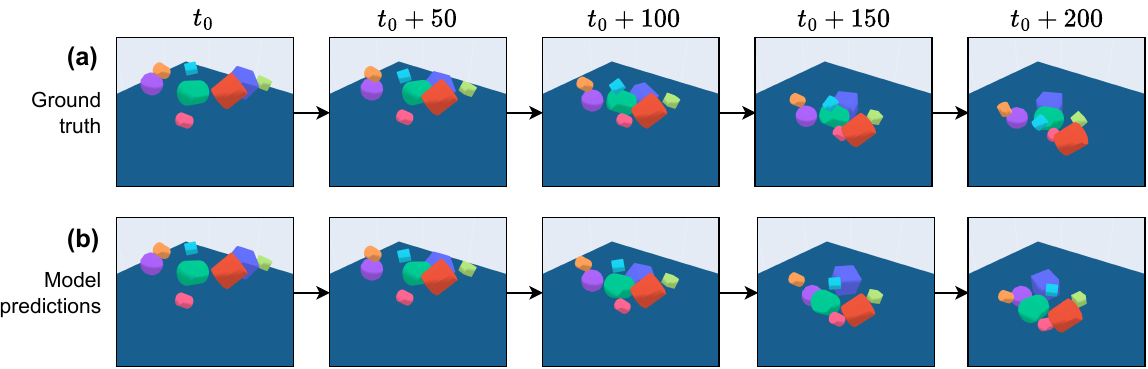}
    \captionsetup{labelfont=bf}
    \caption[Rollout example]{
        \textbf{Rollout example on the \texttt{MOVi-A} dataset.} (\textbf{a}) Ground truth\footnotemark{ }; (\textbf{b}) Model predictions. Our model makes very accurate and realistic predictions on the trajectories of rigid bodies despite complex multi-objects collisions and long rollout durations. The illustrated sequence is taken from experiment 6 of the \texttt{MOVi-A} dataset.
    }
    \label{fig:standard_rollouts}
    \vspace{-0.5cm}
\end{figure}

HOPNet is designed to predict the forward dynamics of all dynamic objects within a scene. During training, HOPNet outputs per-node $\hat{\ddot{\mathbf{x}}}_i$ and per-object accelerations $\hat{\ddot{\mathbf{x}}}_k$, which are compared to ground truth values to optimize the model weights.

During inference, only the per-node accelerations are integrated to compute future node positions using a second-order forward-Euler method with a timestep $\Delta t =1$. Given the previous and current positions $\mathbf{x}_i^{t-1}$ and $\mathbf{x}_i^t$, the estimated next node positions $\hat{\mathbf{x}}_i^{t+1}$ are obtained as follows in Equation \ref{eq:euler_integrator}:

\vspace{-0.3cm}
\begin{equation}
    \label{eq:euler_integrator}
    \hat{\mathbf{x}}_i^{t+1} = \hat{\ddot{\mathbf{x}}}^t_i + 2\mathbf{x}_i^t - \mathbf{x}_i^{t-1}
\end{equation}

\vspace{-0.1cm}
Finally, the object-level positions $\{\hat{\mathbf{x}}^{t+1}_k\}_{k=1...K}$ and quaternionic orientations $\{\hat{\mathbf{q}}^{t+1}_k\}_{k=1...K}$ are determined using shape-matching \cite{shape_matching}, which deterministically computes the optimal object positions and orientations fitting the updated node positions, ensuring consistent rigid body motion. A quaternion $q=(w, x, y, z)$ efficiently represents an object's 3D orientation without ambiguity, unlike Euler angles. More details are available in the "\hyperref[sec:methods]{Methods}" section.\enlargethispage{1\baselineskip}

%%%%%%%%%%%%%%%%%%%%%%%%%%%%%%%%%%%%%%%%%%%%%%%%%%%%%%%%%%%%
\section{Results}

\subsection{Forward dynamics in increasingly complex scenarios}

\begin{figure}[t]
    \centering
    \includegraphics[width=1\textwidth]{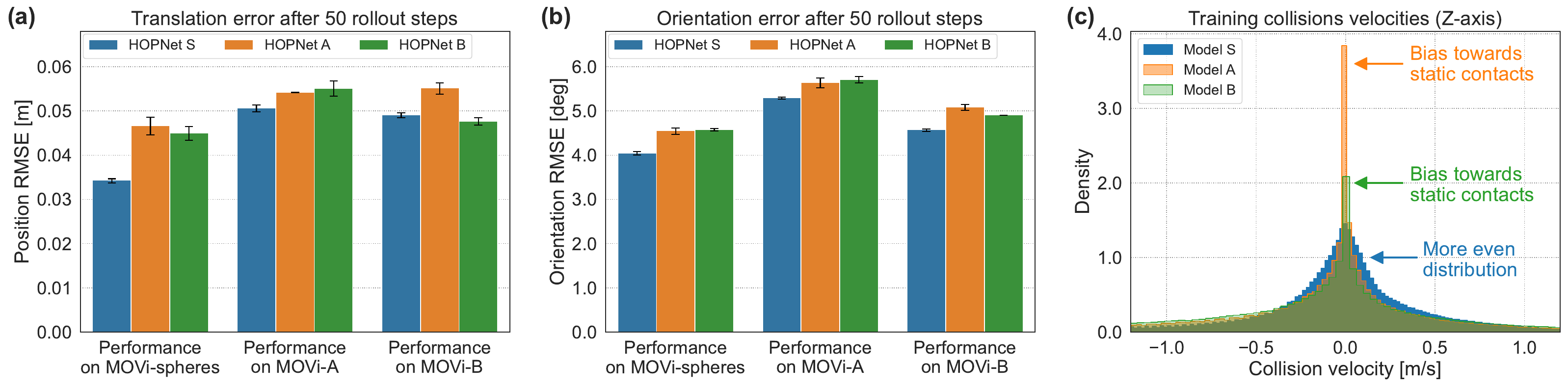}
    \captionsetup{labelfont={color=black,bf}}
    \caption[Generalization results]{
        \textbf{Generalization performance of HOPNet.} (\textbf{a}) Translation and (\textbf{b}) orientation errors after $T=50$ rollout steps of three different models (S, A, and B)  trained on \texttt{MOVi-spheres}, \texttt{MOVi-A}, and \texttt{MOVi-B} respectively. Error bars indicate the mean and standard deviation across three independent random seeds. (\textbf{c}) Distribution of the collision velocities encountered by each model during training. Model S, trained solely on spherical objects, generalizes effectively to more complex object geometries in  \texttt{MOVi-A} and \texttt{MOVi-B}. This ability arises from model S's exposure to a wider range of collision dynamics compared to the slower and more static collisions seen by models A and B. Further details about the datasets are available in Supplementary Information Sections 2 and 3.
    }
    \label{fig:generalization_results}
\end{figure}

To demonstrate the effectiveness of our framework, we evaluate its performance on the task of learning forward dynamics for interacting rigid objects. Specifically, the objective is to accurately learn object interactions and collisions, and to predict the evolution of their positions and orientations.

We conduct our evaluations using four benchmark video datasets of increasing complexity: \texttt{MOVi-spheres}, \texttt{MOVi-A}, \texttt{MOVi-B} and \texttt{MOVi-C} \cite{kubric}. Each experiment consists of a video where a set of rigid objects interact within an environment with unknown characteristics\textemdash such as gravity and air resistance\textemdash that can only be inferred from observations. The \texttt{MOVi-spheres} dataset consists solely of spherical objects, while the \texttt{MOVi-A} includes a mixture of cubes, spheres, and cylinders with coarse mesh resolutions (51 to 64 nodes per object). \texttt{MOVi-B} increases complexity by incorporating real-world everyday objects, such as cones and teapots, with meshes containing up to 1590 nodes. The most challenging dataset, \texttt{MOVi-C}, uses high-resolution 3D-scanned household items from Google's Scanned Objects dataset \cite{downs2022google} averaging 7917 nodes per object and reaching up to 50561 nodes. 

Each experiment contains between 3 and 10 randomly selected objects with varying physical parameters (i.e., mass, friction) and sizes.  In \texttt{MOVi-A} and \texttt{MOVi-B}, objects are initially defined with an average dimension of 1 [m] and then assigned a scaling factor of either 0.7 or 1.4. In contrast, objects in \texttt{MOVi-C} retain their real-world sizes but are further scaled by a randomly sampled factor in the range $[0.75, 3.0]$. Objects are initialized with random positions and velocities and then released onto a flat surface to undergo dynamic interactions. Further details about the datasets are provided in Supplementary Information Section 2.1.

We test our framework in an autoregressive manner: the predicted acceleration $\hat{\ddot{\mathbf{x}}}^t$ at time $t$ is used to compute the next position $\hat{\mathbf{x}}^{t+1}$, which is subsequently fed back as input to predict $\hat{\mathbf{x}}^{t+2}$, as elaborated in Section \ref{subsec:topology_groups}. To evaluate the framework's performance over both short-term and long-term predictions, we consider different rollout horizons $T$ ranging from $25$ to $100$ steps. Our primary results are presented in Figure \ref{fig:standard_results} for a rollout horizon of $T=50$, aligning with the reported results by the current state-of-the-art baseline models FIGNet \cite{fignet} and MeshGraphNet \cite{meshgraphnet}. For a more comprehensive analysis across various horizons $T = \{25,50,75,100\}$, please refer to Supplementary Information Section 3.1.

Due to the absence of official reproducible code, we report results for two variants of FIGNet: the numbers published in the original paper \cite{fignet}, and a reimplementation FIGNet$^{\text{reimpl}}$ trained on exactly the same datasets than HOPNet. The reimplementation is based on publicly available code, which we extended to support the \texttt{MOVi} datasets (see Supplementary Information Section~2.5 for details).

% Each experiment can be broadly divided into three phases: (1) convergence of objects, (2) collisions between objects, and (3) post-collisions divergence. During the convergence phase, objects move under the influence of gravity, with relatively predictable dynamics as they approach the center of the scene. The second phase involves collisions between objects, where their trajectories undergo sudden and highly nonlinear changes. This phase presents the greatest challenge for prediction, as even minimal differences in initial conditions can dramatically alter the outcome\textemdash especially with coarse mesh approximations, where object surfaces are poorly represented and contain pronounced edges between triangles. Finally, in the divergence phase, the objects separate and continue along new trajectories, influenced by the post-collision momentum. With a larger number of objects, many multi-objects collisions happen and the temporal frontiers between those phases get blurry. An example is visible on Fig. \ref{fig:spatio_temporal_ccs}, where objects converge at $t_0$, collide at $\{t_1, t_2\}$, and diverge at $\{t_3,t_4\}$.

The results presented in Figure \ref{fig:standard_results} demonstrate that our framework effectively learns and predicts the dynamics of interacting rigid bodies with remarkable accuracy, closely aligning with the ground truth. In line with previous research studies, we report the average Root Mean Square Error (RMSE) for both object positions (Fig. \hyperref[fig:standard_results]{4a}) and orientations (Fig. \hyperref[fig:standard_results]{4b}) after 50 timesteps.  On the \texttt{MOVi-A} dataset, our framework achieves high fidelity in capturing both the translational and rotational motions of rigid bodies. Moreover, on the more complex \texttt{MOVi-B} dataset, which includes a diverse set of real-world objects, our framework effectively handles collisions, maintaining robust performance despite the increased complexity. Across all datasets, our method consistently outperforms both state-of-the-art methods MeshGraphNet \cite{meshgraphnet} and FIGNet \cite{fignet} by a substantial margin. This trend also holds for the FIGNet$^{\text{reimpl}}$ baseline trained and evaluated on the same data splits as HOPNet, further confirming the effectiveness of our higher-order representations and physics-informed inductive biases.

Importantly, our method is capable of performing $50\%$ more rollout steps before reaching the same translational error. The best baseline reaches an RMSE of $0.115$ [m] at 50 steps on the \texttt{MOVi-A} dataset, whereas our framework reaches this value after 75 steps. This significant improvement allows more accurate predictions further into the future, supporting better and more robust decision-making in dynamic scenarios. For further analysis of extended rollout horizons, please refer to Supplementary Information Section 3.1.

To further validate the robustness and precision of our framework, we present a challenging multi-object rollout example in Figure \ref{fig:standard_rollouts}. In this scenario, the final positions $\{\hat{\mathbf{x}}_k\}_{i=1...K}$ and quaternionic orientations $\{\hat{\mathbf{q}}_k\}_{i=1...K}$ of all $K$ objects closely match the ground truth trajectories, highlighting the framework's precision and robustness in handling complex interactions among multiple entities.

% TODO: Mention the fact that errors are more linked to subtle changes in the initial conditions of the collisions rather than an issue to learn the dynamics of the objects

\subsection{Out-of-distribution generalization to unseen complex geometries}

\begin{figure}[t]
    \centering
    \vspace{-0.4cm}
    \includegraphics[width=1\textwidth]{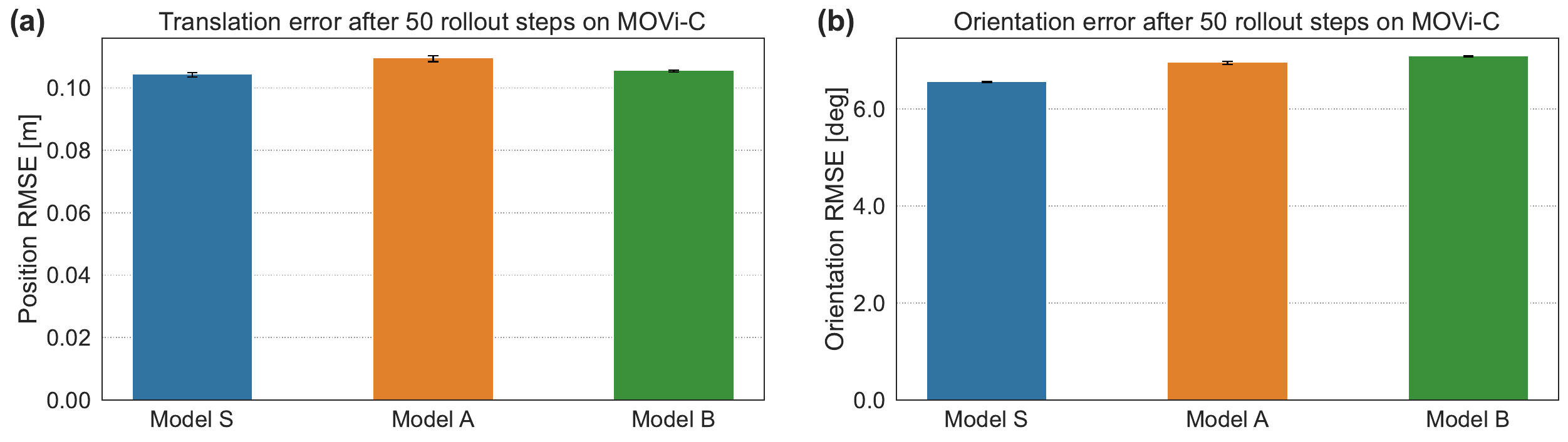}
    \vspace{-0.5cm}
    \captionsetup{labelfont={color=black,bf}}
    \caption[MOVi-C Generalization results]{\textbf{Extreme OOD generalization to complex real-world meshes.} (\textbf{a}) Translation and \linebreak(\textbf{b}) orientation errors after $T=50$ rollout steps on the \texttt{MOVi-C} dataset of three different models (S, A, and B)  trained on \texttt{MOVi-spheres}, \texttt{MOVi-A}, and \texttt{MOVi-B} respectively. Error bars indicate the mean and standard deviation across three independent random seeds.}
    \label{fig:movic_generalization_results}
    \vspace{-0.4cm}
\end{figure}

Beyond achieving high accuracy, HOPNet demonstrates strong Out-Of-Distribution (OOD) generalization to object geometries not encountered during training. This capability arises from the synergy between its physics-informed architecture and higher-order topological representation, which together allow the model to learn generalizable face-face collision dynamics rather than overfitting to object-specific patterns.

We distinguish between two forms of generalization: in-distribution and out-of-distribution (OOD). In-distribution generalization refers to variations within the training distribution, such as different object instances, object counts, or velocity profiles that are already represented in the training data. This type of generalization is typically assessed through standard test-set evaluation and is generally handled well by existing graph-based simulators \cite{battaglia2016interaction}. In contrast, OOD generalization involves more substantial deviations  from the training distribution\textemdash such as entirely novel object geometries, significantly higher object counts, or previously unseen  velocity ranges. Successfully handling these scenarios requires models to capture the underlying physical principles, rather than relying on statistical patterns learned from the training data.

\textbf{OOD generalization to novel geometries} \quad To evaluate HOPNet’s ability to generalize to unseen object geometries, we trained three models (S, A, and B) on the \texttt{MOVi-spheres}, \texttt{MOVi-A}, and \texttt{MOVi-B} datasets, respectively. These datasets span increasing object complexity: S was trained only on spheres, A on basic shapes (including spheres), and B on real-world objects (including basic shapes). All three models were evaluated across all test sets. The results, shown in Figure \ref{fig:generalization_results}, reveal that all three models generalize well, with relatively small differences in performance, even when models are evaluated far outside their training domain.

Notably, model S, trained exclusively on spherical objects, achieves comparable or superior performance on \texttt{MOVi-A} and \texttt{MOVi-B} compared to models A and B (Fig. \hyperref[fig:generalization_results]{7a} and \hyperref[fig:generalization_results]{7b}). This counterintuitive results can be explained by the diversity of collisions in the \texttt{MOVi-spheres} dataset. Because spheres roll and collide continuously, model S is exposed to a rich set of collisions with varying contact angles and speeds. In contrast, objects in \texttt{MOVi-A} and \texttt{MOVi-B} often stabilize on flat surfaces or sharp corners, leading to more redundant interactions (Fig. \hyperref[fig:generalization_results]{7c}). To validate this hypothesis, we modified model B's training by masking slow collisions, reducing the bias toward near-static collisions. The resulting model outperforms the original models B and S, confirming that mitigating the bias toward static contacts leads to better generalization. These findings further support our hypothesis that exposure to diverse, dynamic collisions is critical for learning transferable collision dynamics. Details are provided in Supplementary Information Section 3.3.

\textbf{Extreme OOD generalization to real-world meshes} \quad To further assess HOPNet's OOD generalization, we evaluated all three models on \texttt{MOVi-C}, a dataset featuring high-resolution 3D-scanned objects \cite{downs2022google}. Compared to their respective training sets, \texttt{MOVi-C} scenes contain on average two orders of magnitude more mesh nodes, making this a highly challenging OOD generalization task.

Despite the substantial increase in geometric complexity, all models continue to  perform remarkably well (Fig. \ref{fig:movic_generalization_results}). Consistent with the previous generalization task, model S achieves the lowest errors ($0.104$ [m] translation, $6.564$ [deg] orientation), followed closely by model B ($0.105$ [m], $7.094$ [deg]) and model A ($0.109$ [m], $6.958$ [deg]).

The consistently strong performance across all HOPNet models (S, A, and B) reinforces our hypothesis that learning face-to-face collisions with diverse angles and velocities is sufficient for HOPNet to capture universal collision dynamics, independent of specific geometries or mesh resolutions.

\textbf{Comparison against baselines} \quad Compared to FIGNet \cite{fignet}, HOPNet demonstrates significantly stronger OOD generalization. When evaluated on the "\texttt{MOVi-A} to \texttt{MOVi-B}" scenario, FIGNet shows a $117\%$ increase in translation error (from $0.115$ [m] to $\sim0.250$ [m]) and a $49.8\%$ increase in orientation error (from $14.873$ [deg] to $\sim22$ [deg]) (Appendix G.1 and G.2 of the original FIGNet \cite{fignet} paper). In contrast, HOPNet shows only a $1.85\%$ increase in translation error and even a $9.71\%$ decrease in orientation error. This trend also holds in the more challenging "\texttt{MOVi-B} to \texttt{MOVi-C}" setting, where HOPNet's error increases are $4.0\times$ smaller in translation and $2.7\times$ smaller in orientation than FIGNet's. These results confirm that HOPNet generalizes significantly better to complex, novel object geometries than previous methods.

Additional generalization evaluations comparing HOPNet and FIGNet$^{\text{reimpl}}$ on the same data splits are provided in Supplementary Information Section~3.8.

\begin{figure}[t]
    \centering
    \vspace{-1cm}
    \includegraphics[width=1\textwidth]{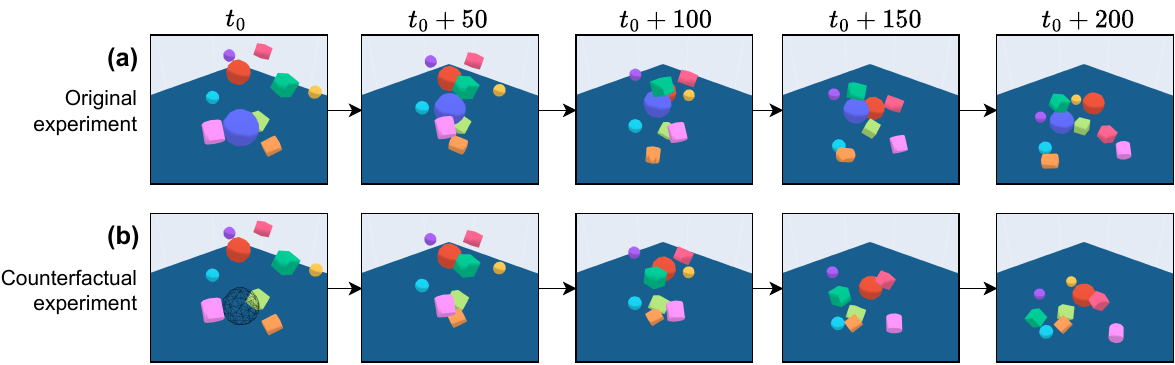}
    \captionsetup{labelfont=bf}
    \caption[Counterfactual rollout example]{
        \textbf{Counterfactual rollout example.} (\textbf{a}) Original experiment\footnotemark{ }; (\textbf{b}) Altered experiment with one sphere removed, shown as a black skeleton to ease understanding. With our dynamic and flexible higher-order topological representation of the environment, our model can easily accommodate any changes in initial conditions: position, orientation, velocity, material properties, ... The illustrated sequence is taken from experiment 21 of the \texttt{MOVi-A} dataset.
    }
    \label{fig:counterfactual_rollout}
    \vspace{-0.5cm}
\end{figure}

\subsection{Counterfactual capabilities}

HOPNet’s accuracy and OOD generalization enable robust counterfactual reasoning, which is essential for planning, analysis, and decision-making. Counterfactual scenarios involve modifying specific scene parameters\textemdash such as object positions, velocities, or the presence of certain elements\textemdash to simulate hypothetical outcomes. While many graph-based simulators support such interventions in principle, achieving accurate long-horizon rollouts under novel conditions remains a significant challenge.\enlargethispage{1\baselineskip}

HOPNet is particularly well-suited for counterfactual analysis thanks to its strong generalization to unseen geometries (Fig. \ref{fig:generalization_results}) and its ability to maintain high predictive accuracy over long rollout horizons. Notably, it exceeds FIGNet by over $50\%$ in rollout length at equivalent error thresholds (Suppl. Sec. 3.1). This capability  is critical, as counterfactual reasoning often requires simulating the long-term effects of initial changes\textemdash where even small prediction errors can quickly compound, undermining the validity of the analysis.

Figure \ref{fig:counterfactual_rollout} illustrates a concrete counterfactual example: a large sphere is removed from the scene at time $t_0$, and HOPNet accurately predicts the resulting evolution of the modified scene. This ability to simulate targeted interventions across diverse object configurations and physical conditions demonstrate the flexibility and practical utility of our approach.

\subsection{Importance of physical consistency}

\begin{figure}[b]
    \centering
    \includegraphics[width=1\textwidth]{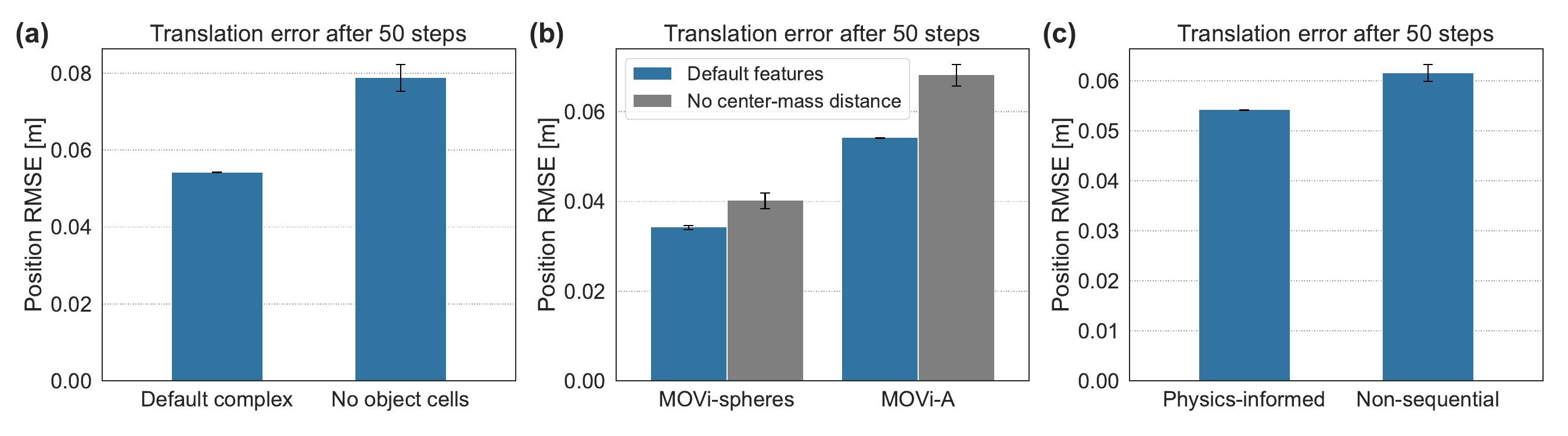}
    \captionsetup{labelfont={color=black,bf}}
    \caption[Importance of physical consistency]{
        \textbf{Importance of physical consistency in HOPNet.}  (\textbf{a}) Ablation study on higher-order topological representations; (\textbf{b}) Ablation study on physics-based features; (\textbf{c}) Ablation study on physics-informed message-passing. Error bars indicate the mean and standard deviation across three independent random seeds.
    }
    \label{fig:ablation_results}
    \vspace{-0.5cm}
\end{figure}

To further demonstrate the impact of our core contributions, we conducted ablation studies on both the higher-order topological representation and the physics-informed inductive biases. For the topological representation, we compared our full higher-order representation against a version that omits the higher-order object-level cells $X^{(4)}$. We did not explicitly test the removal of other higher-order cells, as existing baselines MeshGraphNet \cite{meshgraphnet} and FIGNet \cite{fignet} can be considered as approximations of reduced representations without mesh triangles $X^{(2)}$ and object-level cells $X^{(4)}$, respectively. To assess the importance of physics-informed biases, we performed two key modifications:\linebreak(1) omitting key information inside the node features $\mathbf{h}^{(0)}$, specifically the relative distance between the node and the object's center of mass, and (2) removing the multi-step sequential message-passing. Further architectural details about the models used in these ablation experiments are provided in Supplementary Information Section 3.4.

When comparing our full combinatorial complex representation to its ablated versions, Figure \hyperref[fig:ablation_results]{8a} reveals that removing the object-level cells $X^{(4)}$ leads to a significant drop in performance, with a translation RMSE $44.4\%$ larger than the default HOPNet framework. This decline is expected since intra-object messages are restricted to node-to-node communication, which diminishes the model's expressiveness and fails to capture the rigidity of each object. Although a virtual node is added at the center of each object to enhance multi-hop communication between nodes, the lack of physical consistency results in a noticeable reduction in accuracy. This modified version resembles FIGNet \cite{fignet} in its features but still retains the physics-informed message-passing from HOPNet. With the performance of this ablated model ($\text{RMSE}^{\text{pos}}=0.078$ [m]) falling in between HOPNet ($\text{RMSE}^{\text{pos}}=0.054$ [m]) and FIGNet ($\text{RMSE}^{\text{pos}}=0.115$ [m]), this study also demonstrates the benefits of HOPNet's physics-informed message-passing compared to a standard convolutional one.

Beyond our topological representation, our second key contribution is the incorporation of physics-informed biases within  our topological neural network. To underscore the importance of feature selection, we tested the removal of a critical attribute from the node cell features $\mathbf{h}^{(0)}$: the relative distance between the node and the object's center of mass. This feature is particularly important from a physics standpoint, as it provides structural information about the object and enables the prediction of how collisions near specific nodes affect the object's overall momentum. To evaluate the significance of this feature, we tested our full framework and its simplified version on two datasets: \texttt{MOVi-spheres} and \texttt{MOVi-A}. In the \texttt{MOVi-spheres} dataset, which consists exclusively of spherical objects, removing the "center-mass distance" feature is expected to have a minimal impact, as a sphere's moment of inertia is uniform in all directions. In contrast, the cubes and cylinders in \texttt{MOVi-A} possess anisotropic moments of inertia, making  the nodes' distances to the center of mass critical for accurately predicting collisions with these objects.

 % On the simplest \texttt{MOVi-spheres} dataset, errors are predictably the lowest. When non-spherical objects are introduce in the \texttt{MOVi-A} dataset, the final prediction error increases slightly due to the added complexity of collisions. The anisotropic moments of inertia in cubes and cylinders, along with sharp mesh edges, present challenges for predicting precise collision outcomes.
 
 The results presented in Figure \hyperref[fig:ablation_results]{8b} validate our hypothesis. We only observe a slight performance decline on the \texttt{MOVi-spheres} dataset with $\text{RMSE}^{\text{pos}}$ increasing from $0.034$ [m] to $0.040$ [m], primarily due to the coarse resolution of the spherical meshes. In contrast, there is a substantial drop in accuracy on the \texttt{MOVi-A} dataset, highlighting the critical importance of physics-informed biases for accurately modeling objects with more complex shapes. The $\text{RMSE}^{\text{pos}}$ becomes $26\%$ larger at $0.068$ [m] compared to $0.054$ [m] for the default model. Additionally, on both datasets, the seed-to-seed variability (shown by the error bars) greatly increases compared to the default HOPNet model, underscoring the robustness of our approach.

Finally, we assess the impact of embedding physics-informed inductive biases within our message-passing scheme. Specifically, we compare our physics-informed sequential message-passing (illustrated in Figure \ref{fig:method_overview}) with a non-sequential version, where information flows freely between all cells without any structural guidance. This alternative approach removes the inductive bias that constrains the transfer pathways within our topological neural network. The results, presented in Figure \hyperref[fig:ablation_results]{8c}, clearly demonstrate that without these guided pathways, the model struggles to efficiently learn collision dynamics and has difficulty capturing the interactions among higher-order groups. The $\text{RMSE}^{\text{pos}}$ increases by $14\%$ to $0.061$ [m], even though this variant of the model contains $34\%$ more learnable parameters than the default HOPNet. This highlights the critical role of structured information flow in modeling and learning complex physical interactions.

\subsection{Hyperparameter robustness and efficiency}

Our framework demonstrates significant robustness to variations in hyperparameter selection, highlighting its efficiency and generalizability across diverse applications. When testing a smaller variant of HOPNet by reducing the total number of learnable parameters by $74\%$ through a decreased hidden embedding size, we observed only a minor decrease in performance. This finding confirms the framework's efficiency in learning rigid-body interactions. Additionally, increasing the total number of message-passing steps leads to improved accuracy, a trend consistent with conventional GNNs. A detailed analysis of model size effects is provided in Supplementary Information Section 3.5.

Moreover, our framework shows minimal sensitivity to critical modeling hyperparameters, such as the collision radius $d_c$ used to define collision contact cells $X^{(3)}$. Performance remained stable even when the collision radius was halved or doubled, further validating the method’s stability across a wide range of hyperparameter values. This robustness minimizes the need for extensive fine-tuning, thereby broadening our framework's applicability across different tasks and scenarios. Additional details on this analysis can be found in Supplementary Information Section 3.6.

% TODO: ADD CLEAR NUMBERS FOR INCREASE/DECREASE ONCE AVAILABLE

%%%%%%%%%%%%%%%%%%%%%%%%%%%%%%%%%%%%%%%%%%%%%%%%%%%%%%%%%%%%
\section{Discussion}

% Mention clear contributions (feedback from Zhichao)

In this work, we introduced a novel topology-based framework for modeling rigid bodies and learning their dynamics, addressing critical limitations in existing approaches. By incorporating higher-order topology groups and leveraging physics-informed inductive biases, our framework achieves superior accuracy in predicting complex collisions and interactions between rigid objects. The core strength of our approach lies in its alignment with real-world physics, where we employ domain knowledge to enhance data modeling and design a neural architecture operating on these higher-order structures.

To the best of our knowledge, HOPNet is the first framework to employ higher-order topology groups for modeling spatiotemporal dynamic systems and performing regression instead of classification. Our method demonstrates significant improvements over state-of-the-art baseline models in capturing highly nonlinear dynamics, learning purely from observations. These results highlight the potential of applying higher-order topological representations to other geometric learning tasks, particularly for multi-entity systems governed by known physical laws or structured graph-based data.

% Mention possibility to extend our method (deformable interactions, general force fields, ...)

We validated our approach on a range of increasingly complex datasets, from simple shapes to real-world objects with intricate geometries. In addition to outperforming baseline models, our framework demonstrated robust generalization to unseen scenarios and the ability to address counterfactual questions\textemdash an essential feature for decision-making in dynamic systems. Although this work focuses on rigid-body dynamics, the framework is adaptable enough to extend to deformable objects, such as soft bodies or fluids, and to incorporate additional forces like magnetic or electric fields.

% Discuss minor limitations (need to use knowledge about graph topology, scaling to extremely complex meshes, ...)

While our framework requires additional effort to represent scenes using combinatorial complexes compared to simpler graph-based methods, the resulting enhanced accuracy and learning efficiency underscore the value of this increased complexity. In particular, the integration of physics-informed message-passing enables more accurate predictions than standard GNNs and offers greater robustness to hyperparameter selection. This robustness facilitates the broad generalization of our framework across a wide range of diverse tasks.

Scaling our method to extremely complex meshes with millions of faces remains a challenge, primarily due to the computational cost of message passing over large topological structures. Mesh simplification techniques, such as quadric error decimation\cite{garland1997surface}, can reduce the complexity but often at the cost of losing geometric details.  Adaptive mesh refinement \cite{narain2012adaptive} offers greater  flexibility by dynamically increasing resolution in regions of interest, for example around  collision events. Hierarchical message passing, which has recently  shown promise in fluid simulations \cite{bleeker2025neuralcfd}, could further improve efficiency by enabling multi-scale information propagation. While implicit surface models avoid  explicit meshing altogether, their applicability to dynamic, physics-driven  scenes remains an open question. In future work, we aim explore these directions  to improve scalability without compromising  physical fidelity.

Finally, while our framework currently relies on explicit state information (e.g., geometry, mass, friction) for counterfactual reasoning and generalization, possible extensions could infer these parameters implicitly from data (past observations) in other environments.  Additionally, coupling our model with visual perception systems is a promising direction for further exploration, enabling learning directly from raw pixel data.

% Summary
In summary, our framework effectively models rigid-body dynamics by integrating topological representations with physics-informed neural architectures. Our findings are consistent with recent research that underscores the critical role of higher-order representations in enhancing both the efficiency and accuracy of deep learning models for complex systems. By leveraging these advanced representations, our approach not only aligns with but also advances the current understanding of how deep learning can effectively tackle intricate dynamic interactions.

%TC:ignore
%%%%%%%%%%%%%%%%%%%%%%%%%%%%%%%%%%%%%%%%%%%%%%%%%%%%%%%%%%%%
\section{Methods} \label{sec:methods}

\subsection{Representation of rigid body systems}

% Mention floor representation

We introduce a novel representation for environments comprising $K$ interacting rigid bodies. As previously discussed, current methods typically utilize standard graphs $\mathcal{G} = (\mathcal{V}, \mathcal{E})$, where vertices $\mathcal{V}$ represent nodes and edges $\mathcal{E}$ denote connections between them. In these approaches, the mesh $M$ of each object, defined by its node positions $\{\mathbf{x}_i\}_{i=1..N}$ and faces $\{f_j\}_{j=1..F}$, is integrated  into a global graph $\mathcal{G}$. However, this integration leads to the loss of essential structural information, as standard graphs are unable to explicitly define faces and individual objects. Consequently, the detailed geometries and specific interactions of rigid bodies are not accurately captured, thereby limiting the effectiveness of these conventional graph-based methods.

In contrast, our framework leverages a combinatorial complex $\mathcal{X}^t$ to model the entire environment at a given time $t$. This complex enables us to explicitly define a hierarchy of higher-order node groups, referred to as cells $X^{(r)}$. Each cell of rank $r$ possesses associated features $\mathbf{h}^{(r)}$ and can exchange messages with other groups of the same or different ranks. This higher-order representation enables a more precise encoding of both object structure and interactions, ensuring that no critical geometric or topological information is lost during the modeling process.

Our framework defines five distinct types of cells to comprehensively capture the dynamics and interactions of rigid bodies: nodes $X^{(0)}$, edges $X^{(1)}$, mesh triangles $X^{(2)}$, collision contacts $X^{(3)}$, and objects $X^{(4)}$. The cells $X^{(0)}$ and $X^{(1)}$ correspond  directly to the nodes $\mathcal{V}$ and edges $\mathcal{E}$ of a conventional graph $\mathcal{G} = (\mathcal{V}, \mathcal{E})$. To accurately capture the geometric structure of each object, we introduce $X^{(2)}$ cells to represent individual mesh triangles, and $X^{(4)}$ cells to represent entire objects. Furthermore, to model collisions and enable interactions between individual objects, we incorporate collision contacts $X^{(3)}$. These $X^{(3)}$ cells link two mesh triangles $X^{(2)}$ from different objects when they come within a specified collision radius $d_c$. This combinatorial complex representation $X^t$, alongside the traditional graph representation $\mathcal{G}^t$, is illustrated in Figure \ref{fig:mesh_representation}.

\subsubsection{Node features}
Node features $\mathbf{h}^{(0),\text{features}}$ for node $i$ belonging to object $k$ consist of two key components: (1) finite-difference velocities derived from the node's positional history; and (2) center-mass distances between a node and its object's center of mass. They are formalized in Equation \ref{eq:features_cell0}:

\vspace{-0.3cm}
\begin{equation}
    \label{eq:features_cell0}
    \mathbf{h}^{(0),\text{features}}_i = [\dot{\mathbf{x}}_i^t, \lVert\dot{\mathbf{x}}_i^t\rVert, \dot{\mathbf{x}}_i^{t-1}, \lVert\dot{\mathbf{x}}_i^{t-1}\rVert, \mathbf{d}^{t_0}_i, \lVert\mathbf{d}^{t_0}_i\rVert,\mathbf{d}^t_i, \lVert\mathbf{d}^t_i\rVert]
\end{equation}
\begin{equation*}
    \dot{\mathbf{x}}^t_i = \mathbf{x}^t_i - \mathbf{x}^{t-1}_i \qquad \mathbf{d}_i^t = \mathbf{x}^t_i-\mathbf{x}^t_k
\end{equation*}
where $\dot{\mathbf{x}}^t_i$ and $\dot{\mathbf{x}}^{t-1}_i$ are the finite-difference velocities at time $t$ and $t-1$, $\mathbf{x}_i^t$ is the position of node $i$ at time $t$, $\mathbf{x}_k^t$ is the position of object $k$ at time $t$, $\mathbf{d}^t_i$ and $\mathbf{d}^{t_0}_i$ are the center-mass distances at time time $t$ and $t_0$, $[]$ indicates concatenation and $\lVert\rVert$ is the Euclidean norm. We explicitly include the norm of 3D vectors alongside their components, following prior works \cite{meshgraphnet, fignet}, as it directly relates to physically meaningful quantities like velocity magnitude or force intensity, reducing the need for
intermediate computations and improving the expressiveness of the MLPs.

\subsubsection{Edge features}
Edge features $\mathbf{h}^{(1),\text{features}}_{s\rightarrow r}$ between sender node $s$ and receiver node $r$ include the node-to-node distances measured at time $t_0$ and current time $t$, leading to Equation \ref{eq:features_cell1}:

\vspace{-0.3cm}
\begin{equation}
    \label{eq:features_cell1}
    \mathbf{h}^{(1),\text{features}}_{s\rightarrow r} = [\mathbf{x}^{t_0}_s-\mathbf{x}^{t_0}_r, \lVert\mathbf{x}^{t_0}_s-\mathbf{x}^{t_0}_r\rVert,\mathbf{x}^t_s-\mathbf{x}^t_r, \lVert\mathbf{x}^t_s-\mathbf{x}^t_r\rVert]
\end{equation}
where $\mathbf{x}^t_s$ and $\mathbf{x}^{t_0}_s$ are the positions of sender node $s$ at time $t$ and $t_0$, and $\mathbf{x}^t_r$ and $\mathbf{x}^{t_0}_r$ are the positions of receiver node $r$ at time $t$ and $t_0$.

\subsubsection{Mesh triangle features}
Mesh triangle features $\mathbf{h}^{(2),\text{features}}$ include the normal vector of each mesh triangle $j$:

\vspace{-0.4cm}
\begin{equation}
    \label{eq:features_cell2}
    \mathbf{h}_{j}^{(2),\text{features}} = [\mathbf{n}_{j}^t, \lVert\mathbf{n}_{j}^t\rVert]
\end{equation}
where $\mathbf{n}_{j}^t$ is the normal vector vector of mesh triangle $j$ at time $t$.

\subsubsection{Collision contact features}
Collision contact features $\mathbf{h}^{(3),\text{features}}_{s\rightarrow r}$ between sender triangle $s$ and receiver triangle $r$ are defined by first computing the closest points $\mathbf{p}_s$ and $\mathbf{p}_r$ on each respective triangle. The collision direction is captured by the relative vector $\mathbf{p_s} - \mathbf{p_r}$. Subsequently, a three-axis local coordinate system is established around the closest points on each triangle, utilizing their three vertices $\{s_{i}\}_{{i}=1,2,3}$ and $\{r_{i}\}_{{i}=1,2,3}$. This structured approach, formalized in Equation \ref{eq:features_cell3}, ensures precise modeling of collision dynamics by accurately representing the spatial relationships between interacting mesh triangles:

\vspace{-0.5cm}
\begin{equation}
    \label{eq:features_cell3}
    \mathbf{h}^{(3),\text{features}}_{s\rightarrow r} = [\mathbf{p}_s^t-\mathbf{p}_r^t, \lVert\mathbf{p}_s^t-\mathbf{p}_r^t\rVert, [\mathbf{x}_{s_{i}}^t-\mathbf{p}_s^t,\lVert\mathbf{x}_{s_{i}}^t-\mathbf{p}_s^t\rVert]_{{i}=1,2,3},[\mathbf{x}_{r_{i}}^t-\mathbf{p}_r^t,\lVert\mathbf{x}_{r_{i}}^t-\mathbf{p}_r^t\rVert]_{{i}=1,2,3}]
\end{equation}
where $\mathbf{p}_s$ is the closest point to triangle $r$ on triangle $s$, $\mathbf{p}_r$ is the closest point to triangle $s$ on triangle $r$, $\{\mathbf{x}^t_{s_{i}}\}_{{i}=1,2,3}$ are the positions of all three vertices of triangle $s$ and $\{\mathbf{x}^t_{r_{i}}\}_{{i}=1,2,3}$ are the positions of all three vertices of triangle $r$.

\subsubsection{Object features}
Object features $\mathbf{h}^{(4),\text{features}}$ for object $k$ include finite-difference velocities computed at the object's center of mass. Additionally, a one-hot encoded vector $\mathbf{b}$ distinguishes between static objects and dynamic objects. Finally, some of the object's physical parameters are also included:

\vspace{-0.4cm}
\begin{equation}
    \label{eq:features_cell4}
    \mathbf{h}^{(4),\text{features}}_k = [\dot{\mathbf{x}}_k^t, \lVert\dot{\mathbf{x}}_k^t\rVert, \dot{\mathbf{x}}_k^{t-1}, \lVert\dot{\mathbf{x}}_k^{t-1}\rVert,\mathbf{b},m,c_1,c_2]
\end{equation}
where $\dot{\mathbf{x}}^t_k$ and $\dot{\mathbf{x}}^{t-1}_k$ are the finite-difference velocities at time $t$ and $t-1$, $\mathbf{b}$ is the one-hot encoded object type, $m$ is the object's mass, $c_1$ the friction coefficient and $c_2$ the restitution coefficient.

This combinatorial complex representation $\mathcal{X}^t$ is inherently translation-invariant, as none of the cell features $\mathbf{h}^{(r)}$ include the absolute positions of nodes $\{\mathbf{x}_i\}_{i=1..N}$ or objects $\{\mathbf{x}_k\}_{k=1..K}$. This design ensures a translation-equivariant framework, enhancing the model's generalization capabilities.

HOPNet is a fully autoregressive model: it relies solely on past information from timesteps $t-1$ and $t$ to predict the next state at $t+1$. None of the cell features $\mathbf{h}^{(r),\text{features}}$ in HOPNet incorporate any future information.

\subsection{Physics-informed message passing}

\begin{figure}[t]
    \centering
    \includegraphics[width=1.0\textwidth]{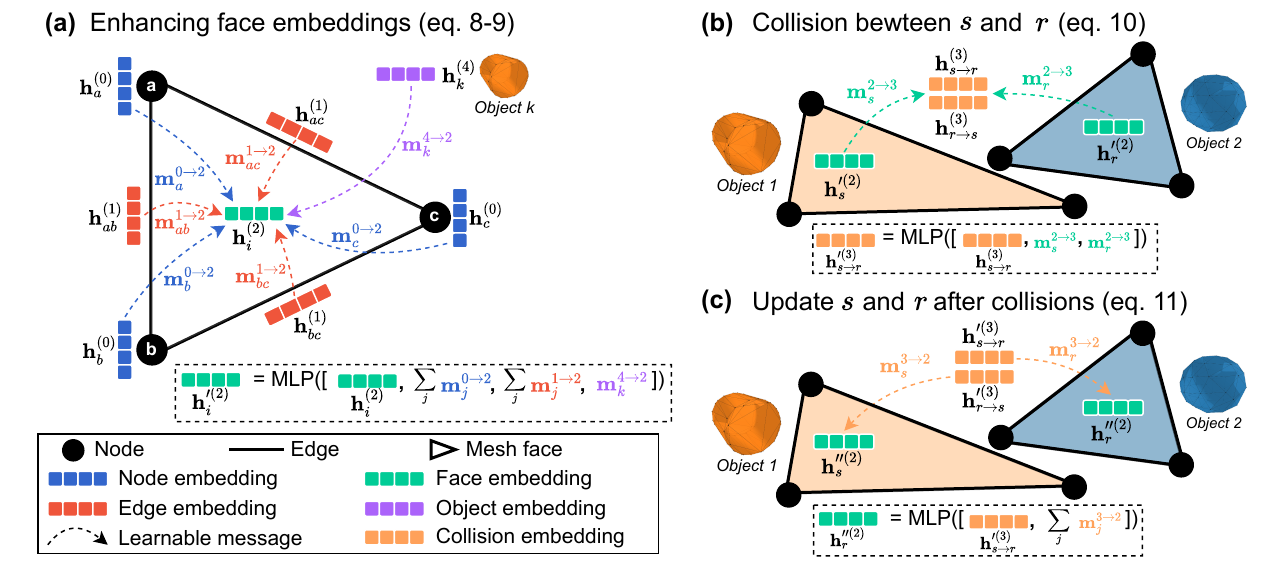}
    \captionsetup{labelfont={color=black,bf}}
    \caption[Physics-informed message passing]{
        \textbf{Physics-informed message passing.}  (\textbf{a}) Enhancing mesh face cells $X^{(2)}$ embeddings with nodes, edges, and object cells; (\textbf{b}) Computing the effect of a collision from two mesh faces $s$ and $r$ on each other; (\textbf{c}) Updating triangles $s$ and $r$ after collisions. Here, $\mathbf{h}^{(r)}$ denotes the embedding of a cell of rank $r$, and $\mathbf{m}^{r_a \rightarrow r_b}$ represents a message passed from a cell of rank $r_a$ to one of rank $r_b$.
    }
    \label{fig:message_passing}
\end{figure}

After constructing a combinatorial complex $\mathcal{X}$ to accurately represent rigid bodies, we design a neural architecture that directly operates on this topological structure. Standard message-passing Graph Neural Networks (GNNs) \cite{gnn} are inadequate for processing higher-order node groups. Instead, we introduce a message-passing Topological Neural Network (TNN) \cite{topological_deep_learning}, which efficiently exchanges information between cells of varying ranks. To ensure consistency and clarity, we adopt the terminology and notations introduced by \cite{topological_deep_learning}, avoiding the need to redefine conventions.

% Introduce how messages are exchanged between cells at a high-level

To introduce how message-passing operates between different $X^{(r)}$ cells, we can compare it to graph-based message-passing, which is a specific case of topological message-passing. In a graph $\mathcal{G}$, a message is sent from a node $v_1$ to another node $v_2$ if an edge $e_{v_1\rightarrow v_2}$ exists. Additionally, if edges also have embeddings, nodes $v_1$ and $v_2$ can send messages to the edge $e_{v_1\rightarrow v_2}$. Hence, in graph message-passing, the edges $\mathcal{E}$ define the neighborhood of nodes and their ability to communicate.

In a combinatorial complex, the neighborhoods $\{\mathcal{N}\}$ of a cell $x^{(r)}$ also define its ability to exchange messages with other cells. Unlike graphs, a cell $x^{(r)}$ has multiple neighborhoods, as neighboring cells can have a lower, same, or higher rank. The $k$-up neighborhood function $\mathcal{N}_{\nearrow,k}(x)$ of a cell $x$ is defined as the set $\{y\in \mathcal{X} | x \subsetneq y, \text{rk}(y)=\text{rk}(x)+k\}$ where $\text{rk}(x)$ is the "rank function", returning the rank $r$ of cell $x^{(r)}$. Similarly, the $k$-down neighborhood function $\mathcal{N}_{\searrow,k}(x)$ of a cell $x$ is defined as the set $\{y\in \mathcal{X} | y \subsetneq x, \text{rk}(y)=\text{rk}(x)-k\}$.

Leveraging neighborhood functions, we construct a message-passing neural architecture operating on the combinatorial complexes $\{\mathcal{X}^t\}$. Our architecture follows an Encode-Process-Decode \cite{graph_network_simulator} paradigm. In the encoding step, the cell features $\mathbf{h}^{(r),\text{features}}$ are passed through separate MLPs, one for each cell rank:

\vspace{-0.4cm}
\begin{equation}
    \mathbf{h}^{(r)} = \text{MLP}^{\text{encoder}}_r(\mathbf{h}^{(r),\text{features}})
    \vspace{-0.1cm}
\end{equation}

% Include explicit equations of the message-passing between cells

The processing step is where sequential message-passing occurs. While the topological representation $\mathcal{X}$ and the neighborhood functions $\mathcal{N}(x)$ offer significant flexibility, theoretically supporting learnable messages between all cell ranks, this approach would introduce unnecessary complexity, computational overhead, and negatively impact the learning efficiency of the network. Therefore, it is essential to carefully constrain message-passing pathways, guiding the information flow to only relevant neighboring cells. To define these pathways, we use physics knowledge about collisions and create a sequential message-passing structure, visible in Figure \hyperref[fig:method_overview]{3b}.

The message-passing used in HOPNet can be divided into three main parts: (A) enhancing mesh faces with intra-object data to prepare for collisions; (B) processing inter-object collisions using mesh faces; (C) updating individual objects and nodes. These three main parts are outlined in Figure \hyperref[fig:method_overview]{3b}.

\subsubsection{Enhancing mesh faces with intra-object data} In this first part, the embeddings of $X^{(2)}$ cells are enhanced with messages from nodes $X^{(0)}$ using $\mathcal{N}_{\searrow,2}(x)$, edges $X^{(1)}$ using $\mathcal{N}_{\searrow,1}(x)$, and object $X^{(4)}$ cells using $\mathcal{N}_{\nearrow,2}(x)$. This first main part integrates critical information into the mesh triangles' embeddings $X^{(2)}$, which are later used to process the inter-object collisions. It is illustrated in Figure \hyperref[fig:message_passing]{9a}, while the messages are formally defined in Equation \ref{eq:step1_msgs}, and the aggregation in Equation \ref{eq:step1_aggr}:

\vspace{-0.5cm}
\begin{gather}
    \label{eq:step1_msgs}
    \mathbf{m}_j^{0\rightarrow2} = \text{MLP}^{\text{processor}}_{m0\rightarrow2}(\mathbf{h}_j^{(0)}) \quad \mathbf{m}_j^{1\rightarrow2} = \text{MLP}^{\text{processor}}_{m1\rightarrow2}(\mathbf{h}_j^{(1)}) \quad \mathbf{m}_j^{4\rightarrow2} = \text{MLP}^{\text{processor}}_{m4\rightarrow2}(\mathbf{h}_j^{(4)})\\
    \label{eq:step1_aggr}
    \mathbf{h}'^{(2)}_i = \text{MLP}^{\text{processor}}_2([\mathbf{h}_i^{(2)},\sum_{\forall j \in \mathcal{N}_{\searrow,2}(x_i^{(2)})}\mathbf{m}_j^{0\rightarrow2},\sum_{\forall j \in \mathcal{N}_{\searrow,1}(x_i^{(2)})}\mathbf{m}_j^{1\rightarrow2},\sum_{\forall j \in \mathcal{N}_{\nearrow,2}(x_i^{(2)})}\mathbf{m}_j^{4\rightarrow2}])
\end{gather}
where $\mathbf{m}^{0\rightarrow2}$, $\mathbf{m}^{1\rightarrow2}$, and $\mathbf{m}^{4\rightarrow2}$ are learnable messages\textemdash sent from nodes, edges, and object cells to mesh triangles respectively\textemdash and $\mathbf{h}'^{(2)}$ are the new mesh triangle embeddings after this message-passing step. 

\subsubsection{Processing inter-object collisions using mesh faces}

This second major part, particularly critical in our framework as it models the inter-object collisions, is presented in Figure \hyperref[fig:message_passing]{9b} and \hyperref[fig:message_passing]{9c}. To ensure energy conservation during each individual collision between two mesh triangles $x_s^{(2)}$ and $x_r^{(2)}$, the messages $m^{2\rightarrow2}_{s\rightarrow r}$ and $m^{2\rightarrow2}_{r\rightarrow s}$ are built in a permutation-invariant manner using the unidirectional collision contacts cells $x^{(3)}_{s\rightarrow r}$ and $x^{(3)}_{r\rightarrow s}$. First, the collision effect from $x^{(2)}_s$ on $x^{(2)}_r$ is computed using $x^{(3)}_{s\rightarrow r}$ (step 2, Eq. \ref{eq:step2}). Then, the new embedding $\mathbf{h}''^{(2)}_{r}$ of the receiver triangle $x^{(2)}_r$ is calculated by aggregating all incoming collision messages on this mesh triangle (step 3, Eq. \ref{eq:step3}), leading to:

\vspace{-0.5cm}
\begin{gather}
    \label{eq:step2}
    \mathbf{m}_{s}^{2\rightarrow3} = \text{MLP}^{\text{processor}}_{m2\rightarrow3}(\mathbf{h}_s'^{(2)}) \qquad \mathbf{h}'^{(3)}_{s\rightarrow r} = \text{MLP}^{\text{processor}}_{3}([\mathbf{h}_{s\rightarrow r}^{(3)},\mathbf{m}_{s}^{2\rightarrow3},\mathbf{m}_{r}^{2\rightarrow3}])\\
    \label{eq:step3}
    \mathbf{m}_{s\rightarrow r}^{3\rightarrow2} = I(\mathbf{h}_{s\rightarrow r}'^{(3)}) \qquad \mathbf{h}''^{(2)}_{r} = \text{MLP}^{\text{processor}}_{2'}([\mathbf{h}_{r}'^{(2)},\sum_{\forall j \in \mathcal{N}_{\nearrow,1}(x^{(2)}_{r})}\mathbf{m}_{j\rightarrow r}^{3\rightarrow2}])
\end{gather}
where $I$ is the identity function, $\mathbf{m}_{s}^{2\rightarrow3}$ and $\mathbf{m}_{r}^{2\rightarrow3}$ are learnable messages from sender triangle $s$ and receiver triangle $r$, $\mathbf{h}'^{(3)}_{s\rightarrow r}$ is the updated collision contact embedding after message-passing step 2, $\mathbf{m}_{s\rightarrow r}^{2\rightarrow3}$ is a learnable message from the collision contact $x^{(3)}_{s\rightarrow r}$ to the receiver triangle $x^{(2)}_{r}$, and $\mathbf{h}''^{(2)}$ are the new mesh triangle embeddings after message-passing step 3.

\subsubsection{Updating intra-objects cells after collisions} This part handles the intra-object updates by integrating the effects of incoming collisions and the current velocity and momentum. The information about incoming collisions is embedded in the updated triangle features $\mathbf{h}''^{(2)}$, while the velocity and momentum at time $t$ are encoded in the node embeddings $\mathbf{h}^{(0)}$ and object embeddings $\mathbf{h}^{(4)}$. In step 4 (Eq. \ref{eq:step4}), updated object embeddings $\mathbf{h}'^{(4)}$ are computed using the incoming collisions. Subsequently, the final object embeddings $\mathbf{h}''^{(4)}$ and node embeddings $\mathbf{h}'^{(0)}$ are derived (Eq. \hyperref[eq:step5]{13-15}), where $k$ is the index of the object containing node $i$, $F$ is the total number of faces in object $k$, and $N$ is the total number of nodes in object $k$.

\vspace{-0.5cm}
\begin{gather}
    \label{eq:step4}
    \mathbf{m}_{i}^{2\rightarrow4} = \text{MLP}^{\text{processor}}_{m2\rightarrow4}(\mathbf{h}_i''^{(2)}) \qquad \mathbf{h}'^{(4)}_{k} = \text{MLP}^{\text{processor}}_{4}([\mathbf{h}_{k}^{(4)},\frac{1}{F}\sum_{\forall j \in \mathcal{N}_{\searrow,2}(x^{(4)}_{k})}\mathbf{m}_{j}^{2\rightarrow4}])\\
    \label{eq:step5}
    \mathbf{m}_{i}^{0\rightarrow4} = \text{MLP}^{\text{processor}}_{m0\rightarrow4}(\mathbf{h}_{i}^{(0)}) \qquad \mathbf{m}_{k}^{4\rightarrow0} = \text{MLP}^{\text{processor}}_{m4\rightarrow0}(\mathbf{h}_{k}'^{(4)})\\
    \mathbf{h}'^{(0)}_{i} = \text{MLP}^{\text{processor}}_{0}([\mathbf{h}_{i}^{(0)},\mathbf{m}_{k}^{4\rightarrow0}])\\
    \label{eq:step5_aggr}
    \mathbf{h}''^{(4)}_{k} = \text{MLP}^{\text{processor}}_{4'}([\mathbf{h}_{k}'^{(4)},\frac{1}{N}\sum_{\forall j \in \mathcal{N}_{\searrow,4}(x^{(4)}_{k})}\mathbf{m}_{j}^{0\rightarrow4}])
\end{gather}
where $\mathbf{m}^{2\rightarrow4}$ are learnable messages from mesh triangles to their parent object, $\mathbf{h}'^{(4)}$ are the new object embeddings after message-passing step 4, $\mathbf{m}^{0\rightarrow4}$ are learnable messages from nodes to object cells, $\mathbf{m}^{4\rightarrow0}$ are learnable messages from object to node cells, $\mathbf{h}'^{(0)}$ the final node embeddings, and $\mathbf{h}''^{(4)}$ the final object embeddings after message passing step 5.

\subsubsection{Predicting accelerations and the next pose} This last part contains the final decoding step of our Encode-Process-Decode architecture (step 6). It simply consists of a readout step on the object embeddings $\mathbf{h}''^{(4)}$ and node embeddings $\mathbf{h}'^{(0)}$ with two MLPs (Eq. \ref{eq:step6}).

\vspace{-0.3cm}
\begin{equation}
    \label{eq:step6}
    \hat{\ddot{\mathbf{x}}}_{i} = \text{MLP}^{\text{decoder}}_{0}(\mathbf{h}_i'^{(0)}) \qquad \hat{\ddot{\mathbf{x}}}_{k} = \text{MLP}^{\text{decoder}}_{4}(\mathbf{h}_{k}''^{(4)})
\end{equation}
where $\hat{\ddot{\mathbf{x}}}_{i}$ and $\hat{\ddot{\mathbf{x}}}_{k}$ are the predicted accelerations for nodes and objects, respectively.

Finally, the last element of our framework consists in predicting the poses of all objects at time $t+1$. It is exclusively applied during inference, not training. Here, we use the second-order forward-Euler integration to predict the next node positions $\hat{\mathbf{x}}^{t+1}_i$. To ensure the objects maintain their rigidity over extended rollouts, we use shape matching \cite{shape_matching}. Using the predicted node positions $\{\hat{\mathbf{x}}_i\}_{i=1..N}$ of each object $k$, we compute the updated center of mass $\hat{\mathbf{x}}^{t+1}_k$ and quaternionic orientation $\hat{\mathbf{q}}^{t+1}_k$ for each object. Finally, we apply the  rotation $\hat{\mathbf{q}}^{t+1}_k$ then translation $\hat{\mathbf{x}}^{t+1}_k$ to the reference mesh $M^t$, and obtain the final node positions predictions.

The choice of aggregation function in HOPNet’s message passing depends on whether the number of contributing entities is fixed or variable. When the count is constant\textemdash as in the case of updating face features using information from nodes, edges, and object (Eq. \ref{eq:step1_aggr})\textemdash we use \texttt{SUM}, enabling the network to learn appropriate scaling factors. In contrast, when the number of messages varies, such as during information exchange between nodes and objects (Eq. \ref{eq:step4} and \ref{eq:step5_aggr}), we use \texttt{MEAN} to maintain  consistency across different mesh resolutions. An exception is made during collision processing (Eq. \ref{eq:step3}), where we apply \texttt{SUM}  despite the varying number of incoming collisions per face,  in order to preserve Newton's law of force summation.

HOPNet's physics-informed message-passing strategy is partially illustrated in Figure \ref{fig:message_passing}, which depicts parts A and B of the message-passing. A comprehensive overview of the full message-passing strategy is provided in Supplementary Figure 3 (Suppl. Sec. 2.2).

\subsection{Performance evaluation metrics}

We use the Root Mean Square Error (RMSE) between the predicted and ground truth object trajectories, on both position and rotation. We report the average RMSE per object in meters (for position) and degrees (for rotation). The exact formulas are given in Equations \ref{eq:rmse_pos} and \ref{eq:rmse_ori}, where $\langle\mathbf{q}\rangle$ is the norm of the vector part of quaternion $\mathbf{q}$ and $\times$ is the Hamilton product.

\vspace{-0.3cm}
\begin{equation}
    \label{eq:rmse_pos}
    \text{RMSE}^{\text{pos}} = \sqrt{\frac{1}{K}\sum_{k=1}^{K}\lVert\hat{\mathbf{x}}-\mathbf{x}\rVert^2}\hspace{-0.5cm}
\end{equation}
\begin{equation}
    \label{eq:rmse_ori}
    \text{RMSE}^{\text{ori}} = \sqrt{\frac{1}{K}\sum_{k=1}^{K}\frac{360}{\pi}\arcsin{(\langle\hat{\mathbf{q}}\times\mathbf{q}^{-1}\rangle)}}\hspace{-0cm}
\end{equation}
\vspace{-0.4cm}

\subsection{Ethics and inclusion statement}

This work introduces a general-purpose, learning-based framework for simulating rigid body dynamics across diverse environments, with applications in fields such as robotics, mechanics, and material science. By learning directly from data, our approach can model physical interactions in complex or partially observed environments, supporting more inclusive and adaptable simulation tools.

The models are trained on synthetic datasets that do not involve human subjects or sensitive attributes. We encourage responsible deployment of this technology, especially in scenarios involving real-world decision-making or safety-critical applications.

\section*{Data availability}

The data generated in this study for training and testing have been deposited in the Zenodo database under accession code \url{https://doi.org/10.5281/zenodo.15800434}. The datasets used in our experiments were generated using the open-source \href{https://github.com/google-research/kubric}{\texttt{Kubric}} library \cite{kubric}. The full datasets can be exactly reproduced using the configuration provided in Supplementary Information Section 2.

\section*{Code availability}

The implementation of the proposed method is based on PyTorch \cite{pytorch} and TopoNetX \cite{toponetx}. Our reproducible code is available at \url{https://github.com/AmauryWEI/hopnet}.

%%%%%%%%%%%%%%%%%%%%%%%%%%%%%%%%%%%%%%%%%%%%%%%%%%%%%%%%%%%%
\bibliographystyle{naturemag}
\bibliography{manual}

\section*{Acknowledgments}

We thank Vinay Sharma and Ismail Nejjar for helpful discussions. This work was supported by the Swiss National Science Foundation (SNSF) Grant Number 200021\_200461.

\section*{Author contributions}

A.W. and O.F. conceptualized the idea, A.W. developed the methodology, A.W. and O.F. conceived the experiments, A.W. conducted the experiments, A.W. and O.F. analyzed the results, and A.W. and O.F. wrote the manuscript.

\section*{Competing interests}

The authors declare no competing interest.

%TC:endignore
\end{document}